\documentclass[10pt,twocolumn,letterpaper]{article}

\usepackage{iccv}
\usepackage{times}
\usepackage{epsfig}
\usepackage{graphicx}
\usepackage{amsmath}
\usepackage{amssymb}

\usepackage[breaklinks=true,bookmarks=false]{hyperref}

\usepackage{xcolor}

\usepackage{enumitem}

\usepackage{url}

\usepackage[normalem]{ulem}
\useunder{\uline}{\ul}{}

\iccvfinalcopy % *** Uncomment this line for the final submission

\def\iccvPaperID{****} % *** Enter the ICCV Paper ID here

\begin{document}

%%%%%%%%% TITLE
\title{Towards a Rigorous Evaluation of XAI Methods on Time Series}

\author{
Udo Schlegel\\
University of Konstanz\\
Konstanz, Germany\\
{\tt\small u.schlegel@uni-konstanz.de}
\and
Hiba Arnout\\
Siemens CT \& TU Munich\\
Munich, Germany\\
{\tt\small hiba.arnout@siemens.com}
\and Mennatallah El-Assady\\
University of Konstanz\\
Konstanz, Germany\\
{\tt\small menna.el-assady@uni-konstanz.de}
\and
Daniela Oelke\\
Siemens CT\\
Munich, Germany\\
{\tt\small daniela.oelke@siemens.com}
\and Daniel A. Keim\\
University of Konstanz\\
Konstanz, Germany\\
{\tt\small keim@uni-konstanz.de}
}

\maketitle
%\thispagestyle{empty}

%%%%%%%%%%%%%%%%%%%%%%%%%%%%%%%%%%%%%%%%%%%%%%%%%%%%%%%%%%%%%%%%%%%%%%%%%%%%%%%%
\begin{abstract}
%%%%%%%%%%%%%%%%%%%%%%%%%%%%%%%%%%%%%%%%%%
% Problem evaluation of black-box on black-box
%%%%%%%%%%%%%%%%%%%%%%%%%%%%%%%%%%%%%%%%%%

% Applications like predictive maintenance or heartbeat anomaly detection have a high demand for accurate and robust machine learning (ML) classifiers.
% But state-of-the-art performance often comes with the cost of high complexity and black-box ML models.
Explainable Artificial Intelligence (XAI) methods are typically deployed to explain and debug black-box machine learning models.
However, most proposed XAI methods are black-boxes themselves and designed for images.
Thus, they rely on visual interpretability to evaluate and prove explanations.
In this work, we apply XAI methods previously used in the image and text-domain on time series.
We present a methodology to test and evaluate various XAI methods on time series by introducing new verification techniques to incorporate the temporal dimension.
We further conduct preliminary experiments to assess the quality of selected XAI method explanations with various verification methods on a range of datasets and inspecting quality metrics on it.
We demonstrate that in our initial experiments, SHAP works robust for all models, but others like DeepLIFT, LRP, and Saliency Maps work better with specific architectures.
\end{abstract}

%%%%%%%%%%%%%%%%%%%%%%%%%%%%%%%%%%%%%%%%%%%%%%%%%%%%%%%%%%%%%%%%%%%%%%%%%%%%%%%%
\section{Introduction}
% WHY interpretability
Due to state-of-the-art performance of Deep Learning (DL) in many domains ranging from autonomous driving~\cite{huval_empirical_2015} to speech assistance~\cite{dahl_context-dependent_2011} and the developing democratization of it, interpretability and explainability of such complex models captured more and more interest.
% Understanding
Agencies such as DARPA introduced the explainable AI (XAI) initiative~\cite{gunning_d._explainable_2016} to promote the research around interpretable Machine Learning (ML) to foster trust into models.
% Laws to explain
Laws like the EU General Data Protection Regulation~\cite{european_union_european_2018} got ratified to force companies to be able to explain the decisions of algorithms to support fairness and privacy and mitigate trust issues of users and costumers.
% WHY XAI
The desiderata of ML systems (fairness, privacy, reliability, trust building~\cite{doshi-velez_towards_2017}) led to a new selection process for models~\cite{rudin_please_2018}.
Depending on the task either interpretable models, such as decision trees~\cite{hidasi_shifttree:_2011}, or new XAI methods, e.g., local interpretable model-agnostic explanations (LIME)~\cite{ribeiro_why_2016}, on top of trained complex models, for instance, Convolutional Neural Networks (CNN) or Recurrent Neural Networks (RNN), are incorporated to guarantee the interpretability demands~\cite{guidotti_survey_2018}.
% Definitions
Due to these new methods for interpretability on a level above a model, we introduce a few new definitions.
In the following, we refer, e.g., LIME as XAI method.
Explainers are defined as an XAI method used on top of a model to get an XAI explanation of the decision making.

%%%%%%%%%% XAI on Time Series
% XAI only on certain domains
Many prominent XAI methods are tailored onto certain input types such as images, e.g., Saliency Maps~\cite{simonyan_deep_2013}, or text, e.g., layer-wise relevance propagation (LRP)~\cite{bach_pixel-wise_2015}.
% Because of benefit from domain
They often benefit of their domain to explain with certain aspects, such as a heatmap on the input~\cite{samek_evaluating_2017}, as they can be used as an overlay by building an abstract feature importance~\cite{guidotti_survey_2018}.
% However sequences/time dimension
However, for instance, videos (sequences of images) and audio have another temporal dimension which is currently omitted by XAI methods.
%Only NLP sequences
Only limited consideration is taken into account for sequence or temporal data, e.g., on XAI method evaluation on natural language processing~\cite{arras_evaluating_2019}.
% No XAI method for time
There is currently only limited work about XAI on time series data such as interpretable decision tress~\cite{hidasi_shifttree:_2011}, calculating prototypes~\cite{gee_explaining_2019} and using attention mechanisms~\cite{hsu_multivariate_2019}.
% Divide not possible
Dividing the hard task of video classifier explanation into time series and image tasks is not possible as there is no good time series solution.
% More and more time-series due to more sensors
However, due to sensors getting cheaper and cheaper, more time-oriented data besides video and audio is generated, and thus, it is important first to test already prominent XAI methods and discover new ones.
% Time series alone important
Analyzing time series further enables to automate more actions, e.g., heartbeat anomaly detection~\cite{chuah_ecg_2007}, solve new tasks, e.g., predictive maintenance~\cite{mobley_introduction_2002}, and predict stock, e.g., stock market forecasting~\cite{kim_financial_2003}.
% Need for XAI on TS
% Further, highlighting the need for XAI in temporal data not only as a step towards videos.

%%%%%%%%%% Evaluation of XAI on Time Series
% Correctness of explanation
To debug and optimize time series prediction models in diverse tasks, not only understanding is essential but also that the XAI explanation is correct itself~\cite{mohseni_survey_2018}.
% Difficult task
Evaluating and verifying these explanations is a difficult task due to raw time series being large and hardly interpretable even by domain experts themselves, and so an evaluation by raw data and explanation inspection is not feasible.
% Lack of domain knowledge needs a quantifiable approach
Due to this lack of connectable domain knowledge, a quantifiable approach is necessary to verify explanations.
% Computer Vision methods to evaluate
Notably, in computer vision exists some work about the evaluation of explanations~\cite{samek_explainable_2017} (e.g., set relevant pixels to zero~\cite{zeiler_visualizing_2014}), which is also possible to use on time series.
% Problem with time
However, these methods omit temporal dependencies by assuming feature independence or only local (short-term) dependency and thus are only limited verifiable on time-oriented data.
% Evaluation methods for XAI time series
Hence, adapted or novel variants of previous methods are needed to evaluate explanations on time series. 

% Contribution
In this work, we show the practical use of various XAI methods on time series and present the first evaluation of selected methods on a variety of real-world benchmark datasets.
Further, we introduce two sequence verification methods and a methodology to evaluate and check XAI explanations on time series automatically.
In preliminary experiments, we show the results of our verification techniques for the selected XAI techniques and their results.
% The main contributions of this work are the following:
% % Contributions
% \setlist{nolistsep}
% \begin{itemize}[noitemsep]
%     \item Two novel verification techniques for time-oriented XAI method explanations 
%     \item A methodology for XAI explanation evaluation and verification on time series
%     \item Preliminary experiments on XAI methods on time series tasks and datasets
% \end{itemize}

%%%%%%%%%%%%%%%%%%%%%%%%%%%%%%%%%%%%%%%%%%%%%%%%%%%%%%%%%%%%%%%%%%%%%%%%%%%%%%%%
\section{Time Series Explanations}

\begin{figure*}[h!t]
    \centering
    \includegraphics[width=\linewidth]{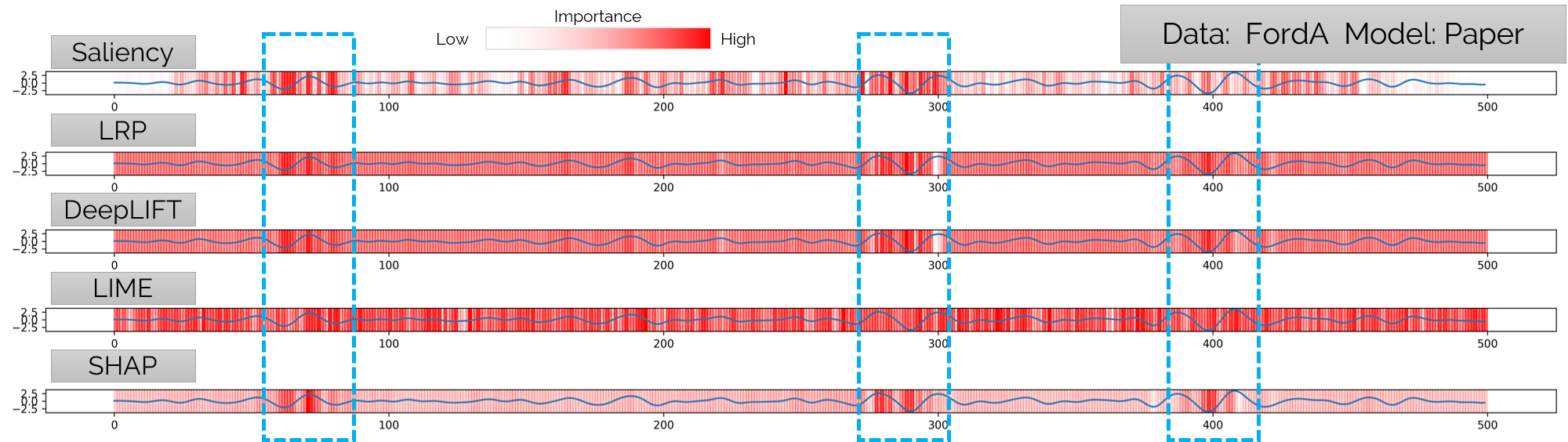}
    \caption{Relevance Heatmaps on an exemplary time series of the FordA dataset using a ResNet paper model.
    XAI methods shown with their relevance heatmaps are Saliency Maps, LRP, DeepLIFT, LIME, and SHAP.
    The blue rectangles display controversial parts of the time series for the XAI methods with red marking high importance for the classification which are, e.g., set to zero by verification methods.}
    \label{fig:saliencymasks}
\end{figure*}

XAI methods have their main application field in computer vision due to the state-of-the-art success of black-box DL models in object recognition and detection~\cite{redmon_yolov3:_2018} and the visual interpretability of the input~\cite{mohseni_survey_2018}.
However, a need for explainability is desired in other domains to either understand the decision making or to improve the models' performance by debugging failures.
Thus, the domain of time series prediction has a high demand for XAI methods.

% Mathematical definitions
A classification dataset with univariate time series data $D$ consists of $n$ samples with classes $c_1, c_2, c_3, ..., c_k$ from a label (multiple classes) with $k$ different classes.
A sample $t$ of $D$ consists of $m$ time points $t = (t_0, t_1, t_2, ..., t_m)$.
E.g., an anomaly detection dataset has only two classes (anomaly, e.g., $c_2$, and normal, e.g., $c_1$).
In the following, the generally considered explanation of most XAI methods is the local feature importance.
Time points get converted to features to introduce a workaround to use XAI methods on time series.
The local feature importance produces a relevance $r_i$ for each time point $t_i$.
Afterward a tuple $(t_i, r_i)$ can be build or more general for the time series vector $t = (t_0, t_1, t_2, ..., t_m)$ a relevance vector can be generated as $r = (r_0, r_1, r_2, ..., r_m)$.

% XAI Method
A model $m$ trained on a subset $X$ from $D$ with labels $Y$ can be formalized to $m(x) = y$ with $x\in X$ and $y\in Y$.
The model $m$ learns based on the provided data $X$, $Y$ to predict an unseen subset $X_{new}$.
In the case of time series, $x$ is a sample like $t = (t_0, t_1, t_2, ..., t_m)$ with $m$ time points.
If then an XAI method $xai$ is incorporated to explain the decisions of such a model, another layer on top of it is created.
An explanation can then be formalized as $xai(x, m) = exp$ with $exp$ being the resulting explanation.
With time series, the explanation $exp$ is a relevance $r = (r_0, r_1, r_2, ..., r_m)$ vector for $m$ time points. 

% Saliency Masks on time series
Similar to the saliency masks on images, a heatmap can be created based on the relevance produced by XAI methods.
It is possible to create a visualization with this heatmap enriching a line plot of the original time series.
Together with domain knowledge, an expert can inspect the produced explanation visualizations to verify the result qualitatively.
\autoref{fig:saliencymasks}. shows an example of relevance heatmaps on time series.
However, as these heatmaps are hard to interpret and a significant challenge to scale to large datasets or long time series, automated verification needs to be applied.

%%%%%%%%%%%%%%%%%%%%%%%%%%%%%%%%%%%%%%%%%%%%%%%%%%%%%%%%%%%%%%%%%%%%%%%%%%%%%%%%
\section{Evaluating Time Series Explanations}
There are various options on how to evaluate and verify XAI explanations automatically.
In computer vision, a common method consists of a perturbation analysis~\cite{zeiler_visualizing_2014}.
This analysis method substitutes a few pixels (e.g., exchange to zero) of an image according to their importance (most or least relevant pixels).
However, because, e.g., a zero could be an indicator for an anomaly in a time series task, the methodology of evaluation of XAI methods for time series needs specialized heuristics.
We present two novel methods suited explicitly for time series by taking the sequence property of the time-oriented data into account.

\subsection{Perturbation on time series}
At first, a perturbation analysis presents preliminary comparison baselines.
The evaluation is based on the assumption that if relevant features (time points) get changed, the performance of an accurate model should decrease massively.
If random time points of the data get changed, the performance should either stagnate or decrease.

\noindent\textbf{Perturbation Analysis --}
The assumption follows the time series $t = (t_0, t_1, t_2, ..., t_m)$ and the relevance produced by the XAI method as $r = (r_0, r_1, r_2, ..., r_m)$ to get a worse result of the quality metric $qm$ for the classifier if combined.
A time point $t_i$ gets gets changed if $r_i$ is larger than a certain threshold $e$, e.g. the $90$th percentile of $r$.
Due to XAI methods have problems with some time-series samples, the threshold leads to only changing a small number of time points.
In the case of time series, the time point $t_i$ is set to zero or the inverse ($max_{t_i} - t_i$) and leads to the new time series samples $t^{zero}$ and $t^{inverse}$.
% , to the max, and to the min - , $t^{max}$, and $t^{min}$

\noindent\textbf{Perturbation Verification --}
To verify the assumption, a random relevance $r_r = (r_0, r_1, r_2, ..., r_m)$ is used for the same procedure.
The number of changed time points, amount of $r_i$ larger than the threshold $e$, is the same as in the case before to set the same prerequisites for the classifier.
This technique creates new time series like the perturbation analysis such as $t_r^{zero}$ and $t_r^{inverse}$.
The assumption to verify the model and the XAI method with the random relevance method follows the schema that the quality metric $qm$ shows e.g. $qm(t)\geq qm(t_r^{zero}) > qm(t^{zero})$ for a model that maximizes $qm$.
% , $t_r^{max}$, and $t_r^{min}$

\subsection{Sequence Evaluation}
To verify that the model and the XAI method also includes time series features such as slopes or minima, we present two novel sequence-dependent methods.
If the assumptions of the perturbation analysis hold, there is still a lack of evaluation of trends or patterns in the time series.
E.g., for the classification, a decrease to zero could be significant, but the perturbation sets the zero to the max as it is essential for the model and so the classification should get worse.
However, if a model learns the general pattern and generalizes good enough to overcome this change, the testing is useless.
Thus to take the inter-dependency of time points into account, a closer look onto the time points itself is crucial.
We propose two new techniques to test and evaluate XAI methods incorporating this hypothesis.

\noindent\textbf{Swap Time Points --}
The first additional method again takes the time series $t = (t_0, t_1, t_2, ..., t_m)$ and the relevance for it $r = (r_0, r_1, r_2, ..., r_m)$.
However, it takes the time points with the relevance over the threshold as the starting point for further changes of the time series.
So, $r_i > e$ describes the start point to extract the sub-sequence $t_{sub} = (t_i, t_{i+1}, ..., t_{i+n_s})$ with length $n_s$.
The sub-sequence then gets reversed to $t_{sub} = (t_{i+n_s}, ..., t_{i+1}, t_i)$ and inserted back into the time series.
Further, in another experiment, the sub-sequence gets set to zero to test the method.
Also, like in the perturbation analysis, the same procedure is done with a random time point positions to verify the time points relevance again.

\noindent\textbf{Mean Time Points --}
Same as the first additional method, the second one also takes into account the time series $t = (t_0, t_1, t_2, ..., t_m)$ and the relevance for it $r = (r_0, r_1, r_2, ..., r_m)$.
Also, it takes the time points with the relevance over the threshold as the starting point for further changes of the time series.
However, instead of swapping the time points, the mean $\mu$ of the sub-sequence $t_{sub} = (t_i, t_{i+1}, ..., t_{i+n_s})$ is taken to exchange the whole sub-sequence to $t_{sub} = (\mu_{t_{sub}}, \mu_{t_{sub}}, ..., \mu_{t_{sub}})$ and inserted back into the time series.
Further, in another experiment, the sub-sequence gets set to zero to test the method.
Also, like in the perturbation analysis, the same procedure is done with a random time point positions to verify the time points relevance again.

\subsection{Methodology}
The methodology to verify an XAI method is conducted in three stages (model training and evaluation, model explanation creation, explanation evaluation, and verification).
\begin{enumerate}
    \item In the first step, a model learns the training data.
Afterward, the trained model predicts the test data and a quality measure (e.g., accuracy) calculates the performance of the result.
    \item In the next step, a selected XAI method creates explanations for every sample of the test data.
Based on the time point relevance by the explanations, the test data gets changed by the evaluation and verification methods mentioned before.
    \item Then, in the last step, each of these newly created test sets gets predicted by the model, and the quality measure is calculated for the comparison.
\end{enumerate}
If the XAI method produces correct explanations, the assumptions $qm(t)\geq qm(t_r^{c}) > qm(t^{c})$ with $qm$ as the quality measure, $t$ the original time series, $t_r^{c}$ the random changed, and $t^{c}$ the relevant changed time series, holds.

% \begin{figure}
%     \centering
%     \includegraphics[width=\linewidth]{content/paper/figures/workflow_new_new_new_new.png}
%     \caption{Workflow for our methodology to evaluate and verify XAI explanations on time series incorporating four steps (model training, model evaluation, model explanation creation, and explanation evaluation and verification).}
%     \label{fig:workflow}
% \end{figure}

\begin{table*}[h!t!]
\resizebox{\textwidth}{!}{%
\begin{tabular}{|l|llll|l|llll|l|llll}
\hline
\textbf{CNN} & \multicolumn{1}{l|}{Zero} & \multicolumn{1}{l|}{Inverse} & \multicolumn{1}{l|}{Swap} & Mean & \textbf{RNN} & \multicolumn{1}{l|}{Zero} & \multicolumn{1}{l|}{Inverse} & \multicolumn{1}{l|}{Swap} & Mean & \textbf{Paper} & \multicolumn{1}{l|}{Zero} & \multicolumn{1}{l|}{Inverse} & \multicolumn{1}{l|}{Swap} & \multicolumn{1}{l|}{Mean} \\ \hline
Saliency & 0.24 & \textbf{0.45} & 0.39 & 0.34 & Saliency & {\ul \textbf{0.29}} & {\ul \textbf{0.42}} & {\ul \textbf{0.23}} & 0.22 & Saliency & 0.06 & 0.08 & 0.07 & 0.07 \\ \cline{1-1} \cline{6-6} \cline{11-11}
LRP & \textbf{0.44} & 0.39 & {\ul \textbf{0.41}} & {\ul \textbf{0.41}} & LRP & 0.21 & 0.21 & 0.14 & 0.13 & LRP & {\ul \textbf{0.29}} & 0.29 & \textbf{0.29} & 0.34 \\ \cline{1-1} \cline{6-6} \cline{11-11}
DeepLIFT & {\ul \textbf{0.48}} & \textbf{0.45} & \textbf{0.40} & \textbf{0.39} & DeepLIFT & 0.00 & 0.00 & 0.00 & 0.00 & DeepLIFT & {\ul \textbf{0.29}} & \textbf{0.30} & \textbf{0.29} & \textbf{0.35} \\ \cline{1-1} \cline{6-6} \cline{11-11}
LIME & 0.16 & 0.32 & 0.17 & 0.17 & LIME & 0.10 & 0.21 & 0.06 & 0.07 & LIME & 0.02 & 0.06 & 0.04 & 0.02 \\ \cline{1-1} \cline{6-6} \cline{11-11}
SHAP & 0.25 & {\ul \textbf{0.46}} & 0.33 & 0.29 & SHAP & \textbf{0.26} & \textbf{0.35} & {\ul \textbf{0.23}} & {\ul \textbf{0.23}} & SHAP & {\ul \textbf{0.29}} & {\ul \textbf{0.40}} & {\ul \textbf{0.31}} & {\ul \textbf{0.38}} \\ \hline
\textit{Random} & \textit{0.17} & \textit{\textbf{0.45}} & \textit{0.15} & \textit{0.10} & \textit{Random} & \textit{0.13} & \textit{0.23} & \textit{0.03} & \textit{0.03} & \textit{Random} & \textit{0.13} & \textit{0.21} & \textit{0.07} & \textit{0.04} \\ \cline{1-1} \cline{6-6} \cline{11-11}
\end{tabular}%
}
\caption{Results table with the averaged changed accuracy of the different models over all datasets. Change to test accuracy is calculated by normalizing the base accuracy to the one from the changed data.}
\label{tab:results}
\end{table*}
%%%%%%%%%%%%%%%%%%%%%%%%%%%%%%%%%%%%%%%%%%%%%%%%%%%%%%%%%%%%%%%%%%%%%%%%%%%%%%%%
\section{Discussion}
The discussion divides into three parts.
At first, the datasets and employed models are addressed to help to reproduce the experiments.
Afterward, the selected XAI methods are introduced in short, giving an overview.
Lastly, we discuss the preliminary evaluation results.

\subsection{Datasets \& Models}
Nine datasets of the UCR Time Series Classification Archive~\cite{dau_ucr_2018} and a ECG hearbeat dataset~\cite{goldberger_physiobank_2000} are included in a real-world focused preliminary experiment.
These ten datasets, namely FordA, FordB, ElectricDevices, MelbournePedestrian, ChlorineConcentration, Earthquakes, NonInvasiveFetalECGThorax1, NonInvasiveFetalECGThorax2, Strawberry~\cite{dau_ucr_2018}, and Physionet's MIT-BIH Arrhythmia~\cite{goldberger_physiobank_2000}, consist of two different tasks, binary and multi-class prediction.
Primarily, binary classification, e.g., for anomaly detection, is a critical use case for time-series predictions to tackle applications like predictive maintenance or heartbeat categorization.

During the experiments, two different architectures (CNN and RNN) are used as baseline models.
If available, the architecture provided by the dataset paper is also incorporated.
The considered CNN consists of a 1D convolution layer with kernel and channel size of three.
Afterward, a dense layer with 100 neurons learns the classification for a specific problem.
The considered RNN consists of an LSTM layer with 100 neurons and again a dense layer with 100 neurons for the classifier.
Both networks train each dataset individually for 50 epochs.
The paper models consist of ResNet-based architectures and also 50 epochs.

\subsection{XAI methods}
The experiment is conducted with the five most prominent XAI methods (LIME~\cite{ribeiro_why_2016}, LRP~\cite{bach_pixel-wise_2015}, DeepLIFT~\cite{shrikumar_learning_2017}, Saliency Maps~\cite{simonyan_deep_2013}, SHAP~\cite{lundberg_unified_2017}).
LIME employs a so-called surrogate model to explain the decision of an ML model.
Thru sampling data points around an example to be explained, it learns a linear model to extract local feature importance for the prediction of the more complex model.
By propagating the gradients through the network, Saliency Maps and DeepLIFT build a heatmap as feature importance.
LRP propagates a relevance score backward through the underlying model to specify feature importance.
SHAP employs shapely values and game theory to find the best fitting feature to gain the most for the prediction.

\subsection{Results}
Our preliminary results, see \autoref{tab:results}., show that DeepLIFT and LRP have the largest overall quality metric decrease in CNNS for the perturbation and sequence analysis, which shows the working local feature importance.
Saliency Maps and SHAP outperform the others in RNNs by showing quality metric decreases, which is somewhat unexpected but shows a need for further exploration of RNNs with XAI methods.
In more advanced ResNet architectures, SHAP produces the best results.
However, also DeepLIFT and LRP show good results, which again shows the practical local feature importance.
LIME shows terrible results in all cases, most likely due to large dimensionality and the employed linear classifier.
Further, the results show the desiderata for the sequence verification methods as the random perturbation of time points has a significant quality metric decrease.
Our proposed sequence verification methods present more clearly that the assumption $qm(t)\geq qm(t_{random}^{mean}) > qm(t^{mean})$ with $qm$ as the quality measure, $t$ the time series, $t_{random}^{mean}$ and $t^{mean}$ the changed time series holds.

%%%%%%%%%%%%%%%%%%%%%%%%%%%%%%%%%%%%%%%%%%%%%%%%%%%%%%%%%%%%%%%%%%%%%%%%%%%%%%%%
\section{Conclusion and Future Work}
Our methodology and verification methods show that XAI methods, proposed for images and text, work on time series data by specifying a relevance to time points.
The methods also demonstrate that the models take the temporal aspect into account in some cases.
In our experiment, we find that SHAP works robust for all models, but others like DeepLIFT, LRP, and Saliency Maps work better for specific architectures.
LIME performs worst most likely because of the large dimensionality by converting time to features.
However, we also conclude that a demand is given to introduce more suitable XAI methods on time series to guarantee a better human understanding in the process of XAI.
As seen by the hard to interpret visual saliency masks (heatmaps) on time series, a need for a more abstract representation is necessary and increases the importance for more sophisticated visual XAI methods on time series.

{\small
\bibliographystyle{ieee}
\bibliography{lib}
}

\end{document}